\documentclass[10pt,twocolumn,letterpaper]{article} 

\usepackage{avss}
\usepackage{times}
\usepackage{epsfig}
\usepackage{graphicx}
\usepackage{amsmath}
\usepackage{amssymb}
\usepackage{xcolor}
\usepackage{nopageno}
\makeatletter
\@namedef{ver@everyshi.sty}{}
\makeatother
\usepackage{tikz}
\usepackage{pgfplots}
\usepackage{setspace}
\usepackage{subcaption}
\usepackage{caption}
\avssfinalcopy % *** Uncomment this line for the final submission

\def\avssPaperID{12} % *** Enter the AVSS Paper ID here

\ifavssfinal\fi
\begin{document}
\captionsetup[figure]{font=small}

%%%%%%%%% TITLE
\title{Online Multi-Object Tracking with Historical Appearance Matching and \\Scene Adaptive Detection Filtering}

\author{Young-chul Yoon\quad Abhijeet Boragule \quad Young-min Song \quad Kwangjin Yoon \quad Moongu Jeon\\
Gwangju Institute of Science and Technology\\
123 Cheomdangwagi-ro, Buk-gu, Gwangju, 61005, South Korea\\
%123 Cheomdangwagi-ro, Buk-gu, Gwangju, 61005, South Korea\\
{\tt\small \{zerometal9268, abhijeet, sym, yoon28, mgjeon\}@gist.ac.kr}
}

\maketitle
% \thispagestyle{empty}

%%%%%%%%% ABSTRACT
\begin{abstract}
   In this paper, we propose the methods to handle temporal errors during multi-object tracking. Temporal error occurs when objects are occluded or noisy detections appear near the object. In those situations, tracking may fail and various errors like drift or ID-switching occur. It is hard to overcome temporal errors only by using motion and shape information. So, we propose the historical appearance matching method and joint-input siamese network which was trained by 2-step process. It can prevent tracking failures although objects are temporally occluded or last matching information is unreliable. We also provide useful technique to remove noisy detections effectively according to scene condition. Tracking performance, especially identity consistency, is highly improved by attaching our methods.
\end{abstract}

%%%%%%%%% BODY TEXT
\section{Introduction}
Current paradigm of multi-object tracking is tracking by detection approach. Most trackers assume that detections are already given and focus on labeling each detection with specific ID. This labeling process is basically done by data association. For online tracker, the data association problem could be simplified to the bipartite matching problem and the hungarian algorithm has frequently been adopted to solve it. Before solving the data association problem, a cost matrix has to be defined. Each element of the cost matrix is the measure of affinity(similarity) between specific object and detection(observation). Because the data association simply finds 1-to-1 matches on cost matrix, it is important to derive accurate affinity scores for better performance. 

Motion is a basic factor of affinity. Motion is the only information that we can guess in a simple tracking environment(e.g. tracking dots, which are signals from specific objects like ship or airplane, on 2D field). The kalman filter has frequently been adopted for motion modeling. It can model temporal errors by adaptively predicting and updating the positions of objects according to tracking condition. But it is insufficient to track objects in more complex situation. Scenes taken directly from RGB camera contain a lot of difficulties. As described in Figure 1(temporal occlusion), objects are occluded by other objects and obstacles which exist on the scene. To overcome this, we can exploit appearance information. There have been many works \cite{Boragule17, Bae18, Keritz16, Taixe16, Kim2015} which tried to derive accurate appearance affinity. Several works \cite{Boragule17, Keritz16} tried to design appearance model without using deep neural network(DNN). Those trackers achieved better performance but couldn't significantly improve the performance. Along rapid development of deep learning, several works \cite{Bae18, Taixe16, Kim2015} tried to apply DNN to calculating appearance affinity. Most of those works \cite{Bae18, Taixe16} used the siamese network to calculate the affinity score. Although siamese network has a strong discriminating power, it can only see cropped patches which contain limited information. If imperfect detectors \cite{Dollar14,Felzenszwalb10} are used to extract detections, detections themselves contain inaccurate information. Those detections are ambiguous as described in Figure 1(noisy detections) and may lead to inaccurate appearance affinity.

\begin{figure}[t]
\scriptsize
\begin{center}
   \includegraphics[width=0.9\linewidth]{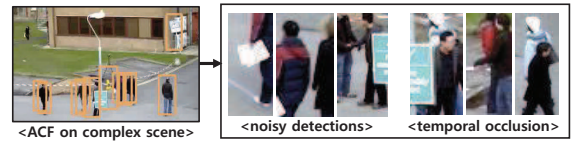}
\end{center}
\vspace{-5mm}
   \caption{Example of temporal errors. ACF detector creates noisy detections which include several objects simultaneously or only a small part of object. Also, a lot of temporal occlusions occur because of complex scene condition}
\label{fig:long}
\label{fig:onecol}
\end{figure}

\begin{figure*}[t]
\begin{center}
   \includegraphics[width=0.9\linewidth]{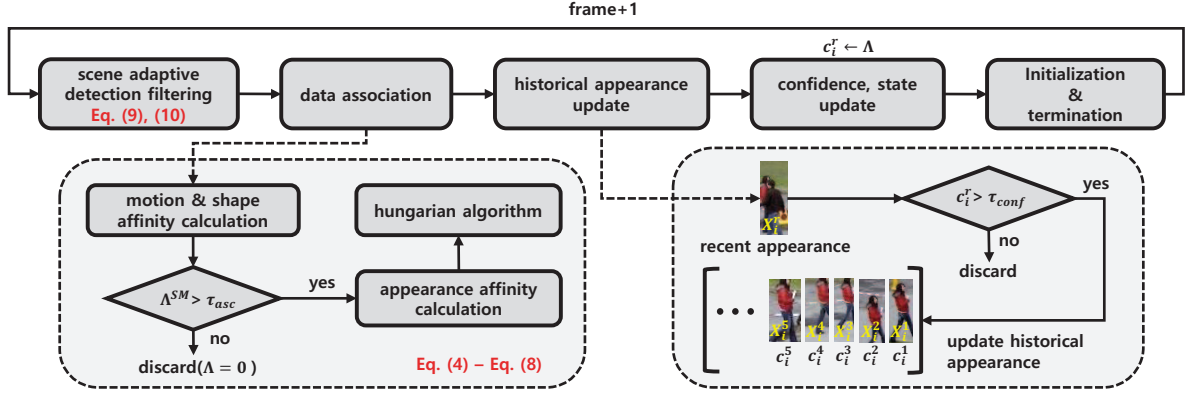}
\end{center}
\vspace{-5mm}
   \caption{Our tracking framework.}
\label{fig:long}
\label{fig:onecol}
\end{figure*}

We propose several methods to tackle aforementioned problems(noisy detections, temporal occlusion). First, it is hard to match a object to an observation when the recent object appearance is ambiguous. To break this, we save reliable historical appearances. From the support of reliable historical appearances, we can get an accurate affinity score even in ambiguous situation. And it is necessary to reduce noisy detections as much as possible for better performance. Many trackers have used a constant detection threshold(e.g. 30) for all sequences. Instead of using a constant threshold, we propose a method to decide the detection threshold according to a scene condition. In our best knowledge, this is the first work that considers to filter out detections according to a scene condition. In summary, our main contributions are:\\

\begin{tabular}{@{\textbullet~}l@{\ }p{3in}}
  &The proposed historical appearance matching method solves the matching ambiguity problem;\\[2mm]
  &The proposed 2-step training method with the siamese network produces the better tracking performance than training by single dataset;\\[2mm]
  &We propose a simple method to adaptively decide detection confidence threshold. It decides the threshold according to scene conditions and performs better than the constant threshold;\\[2mm]
\end{tabular}
In the experiments section, it is proved that each of our method improves the tracking performance.

\section{Proposed methods}
In this section, we describe our tracking framework and proposed methods. Our framework is shown in Figure 2, and is based on the simple framework of the online multi-object tracking. It associates existing objects with observations first. Then, update the states of objects using associated observation and process birth\&death of objects. Our main contribution is to designing appearance cues and preprocessing given detections. It would be explained in following sub-sections. 

\subsection{Affinity models}
Our affinity model consists of three cues: appearance, shape and motion. The affinity matrix is calculated by multiplying scores from each cue: 
\begin{equation}
\label{eq:eg}
\Lambda(i, j)=\Lambda^A(X_i,Z_j)\Lambda^S(X_i,Z_j)\Lambda^M(X_i,Z_j)
\end{equation}
Each of $A, S, M$ indicates appearance, shape and motion. Score from each cue is calculated as below : 
\begin{equation}
\begin{aligned}
\Lambda^S(X,Z) &=\exp\text{\footnotesize{$\Big(-\xi\Big\{\frac{\left|h_{\hat{X}}-h_{Z}\right|}{h_{\hat{X}}+h_Z}+\frac{\left|w_{\hat{X}}-w_Z\right|}{w_{\hat{X}}+w_Z}\Big\}\Big)$}},\\
\Lambda^M(X,Z) &=\exp(-\eta(p_{Z}-p_{\hat{X}})^T\Sigma^{-1}(p_{Z}-p_{\hat{X}})) \\
\end{aligned}
\end{equation}
The appearance affinity score is calculated by our proposed method. It will be explained in section 2.2 and 2.3. Different from other tracking methods, we predict the state of each object $X$ not only for motion but also for shape. Although we model motion and appearance affinities robust to error, tracking may fail because of noisy detections with the different size. We thought the kalman filter could be applied to the shape state $(w, h)$ in a similar way to predict the motion state $p$. $\hat{X}$ indicates the predicted state of the object $X$. We calculate the relative difference of height and width between the object and the observation. Motion affinity score is calculated by the mahalanobis distance between the position of the predicted state and the observation with the predefined covariance matrix $\Sigma$ which generally works well in any scene condition. 
\subsection{Joint-input siamese network}
There are various structures of the siamese network that we can consider to use in multi-object tracking. From experiments of the prior works \cite{Taixe16}, the joint-input siamese network outperforms other types of the siamese network. Also, it is important to set the output range between 0-1 to balance with other affinities(motion and shape). The softmax layer of the joint-input siamese network naturally set the output range between 0-1. Our network structure is described in Figure 3. Different from the prior works, we used batch normalization \cite{Ioffe15} for better accuracy. It prevents overfitting and improves convergence so is useful to train the network with the small size of training data. Thanks to the convolutional neural network which can extract rich appearance features, ours, without historical matching, outperforms the color histogram based tracking(Figure 7(a)). We trained our network in two steps: pretrain and domain adaptation. The detail of the network training is explained in Figure 6 and Section 3.1. 

\begin{figure}[t]
\scriptsize
 \begin{center}
\begin{tabular}{|l|l|l|l|{c}r}
\hline
\multicolumn{1}{|c|}{\textbf{layer}}&\multicolumn{1}{c|}{\textbf{filter size}}&\multicolumn{1}{c|}{\textbf{input}}&\multicolumn{1}{c|}{\textbf{output}}\\
\hline\hline
conv \& bn \& relu &9x9x12 & 128x64x6 & 120x56x12\\
\hline
max pool &2x2 & 120x56x12 & 60x28x12\\  
\hline
conv \& bn \& relu & 5x5x16 & 60x28x12 & 56x24x16\\
\hline
max pool &2x2 & 56x24x16 & 28x12x16\\  
\hline
conv \& bn \& relu & 5x5x24 & 28x12x16 & 24x8x24\\
\hline
max pool &2x2 & 24x8x24 & 12x4x24\\  
\hline
flatten & - & 12x4x24 & 1x1152\\
\hline
dense & - & 1x1152 & 1x150\\  
\hline
dense & - & 1x150 & 1x2\\  
\hline
softmax & - & 1x2 & 1x2\\  
\hline
\end{tabular}
\end{center}
\vspace{-5mm}
\caption{Our joint-input siamese network structure. bn indicates batch normalization layer. Each of two final output means probability of which two inputs are identical or different.}
\end{figure}

\subsection{Historical Appearance Matching}
Because of occlusion and inaccurate detection, the object state may be unreliable. As we mentioned in Section 2.1, the shape and motion cues can handle temporal errors using the kalman filter. However, different from those cues, the size of appearance feature is huge and is hard to be modeled considering temporal errors. Before explaining our method, we revisit the method of the adaptive color histogram update. It is possible to update object color histogram adaptively according to the current matching affinity score. It can be represented as:
\begin{equation}
\label{eq:eg}
Hist^X_t=\alpha Hist^{\hat{X}}_t+(1-\alpha)Hist^X_{t-1}
\end{equation}
$Hist^X_t$ means the saved color histogram of object $X$ in time $t$ and $Hist^{\hat{X}}_t$ is the matched observation color histogram of object $X$ in time $t$. The update ratio is controlled easily by $\alpha$. $\alpha$ is large if the current matching affinity score is high and vice versa. However, the color histogram is still sensitive to change of light, background and object pose. The joint-input siamese network produces much reliable affinity score. But features can't be updated adaptively like the color histogram because input images are concatenated and jointly inferred through network. So, we propose the historical appearance matching(HAM) as following equations:

\begin{figure}[t]
\begin{center}
   \includegraphics[width=0.9\linewidth]{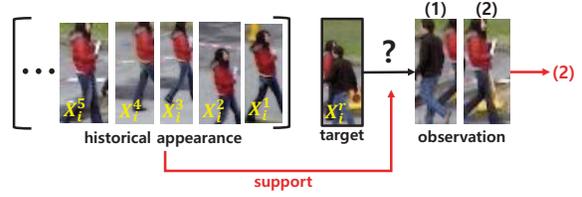}
\end{center}
\vspace{-5mm}
   \caption{Example of breaking ambiguity using historical appearance matching. (Black arrow): It is hard to choose the observation correspond to object among (1) and (2) because of ambiguous recent object appearance. (Red arrow): From support of historical appearance, object can be matched to correct observation (2).}
\label{fig:long}
\label{fig:onecol}
\end{figure}

\iffalse
\begin{equation}
\Lambda^A(X_i,Z_j) = 
\begin{cases}
ham(X_i,Z_j), if \enspace (\exists i' \lor \exists j') \\\hspace{50pt} \{i'|\Lambda^{SM}(i',j) > \tau_{asc}, i'\ne{i}\} \\ \hspace{50pt}\{j'|\Lambda^{SM}(i,j') > \tau_{asc}, j'\ne{j}\}
\\
siam(X_i,Z_j), else
\end{cases}
\end{equation}
\fi
\begin{equation}
\Lambda^A(X_i,Z_j) = 
\begin{cases}
siam(X^r_i,Z_j), &baseline \\
ham(X_i,Z_j), &proposed
\end{cases}
\end{equation}
The baseline method was additionally defined for comparison. $X_i$ and $Z_j$ are the $i$th object and the $j$th observation respectively. Baseline($siam$) takes score from the siamese network with two inputs: $X^r_i$($r$ indicates recently matched appearance) and $Z_j$. And $ham$ is the proposed method that we designed to break confusion and ambiguity as described in Figure 4. $ham$ is calculated as:
\begin{equation}
\begin{aligned}
ham(X_i,Z_j) &= c^{r}_i*siam(X^{r}_i, Z_j) +\\ &(1-c^{r}_i)\sum^{N^h_i}_{n=1}(w^n_i*siam(X^n_i,Z_j))
%&w^n_i = \frac{c^n_i}{\sum^{N^h_i}_{k=1}c^k_i}
\end{aligned}
\end{equation}
where $c^r_i$ is the recent matching confidence(affinity) of the object. Relative weights of two terms in the equation are controlled by $c^r_i$. If the recent matching is unreliable($c^r_i\downarrow$), the second term(reliable historical appearances) takes bigger portion in the appearance affinity and vice versa. $N^h_i$ is a number of the saved historical appearances of the object $X_i$. $w^n_i$ is the relative weight of $X^n_i$($n$-th historical appearance of object $X_i$). It is defined as:
\begin{equation}
\begin{aligned}
&w^n_i = \frac{c^n_i}{\sum^{N^h_i}_{k=1}c^k_i}
\end{aligned}
\end{equation}
Each weight of historical appearance is calculated by dividing its matching confidence($c^n_i$) by sum of all matching confidences. $c^k_i$ indicates the affinity score($\Lambda$) of $X_i$ at the time that the $k$-th historical appearance is matched with $X_i$. So, the sum of all $w^n_i$ becomes 1. And naturally, this assures that $ham$ is in range 0-1. As described in Figure 2, historical appearance of each object is updated when the matching appearance is bigger than $\tau_{conf}(0.6)$. We maintain the maximum number of historical appearances to be not more than 10 and the oldest one to be within 15 from the current frame.

\subsection{Gating Technique}
Processing the siamese network could be a time bottleneck. To reduce processing time, we applied a simple but efficient gating technique from our previous work \cite{YCYoon18}. Before calculating the appearance affinity, we create shape and motion affinity matrices using Eq. 2:
\begin{equation}
\begin{aligned}
&\Lambda^{SM}(i,j) = \Lambda^{S}(X_i,Z_j)\Lambda^{M}(X_i,Z_j),\\
&\text{\scriptsize{$\forall i\in\{1,\cdots,N_X\}, \forall j\in\{1,\cdots,N_Z\}$}}
\end{aligned}
\end{equation}
where $N$ is the total number of object($X$) or observation($Z$). Then, we calculate the final affinity matrix as: 
\begin{equation}
\Lambda(i,j) = 
\begin{cases}
\Lambda^{SM}(i,j)\Lambda^{A}(X_i,Z_j), & \mbox{if } \Lambda^{SM}(i,j) > \tau_{asc} \\ 
0, & else
\end{cases}
\end{equation}
The final affinity matrix is simply calculated by multiplying appearance affinities when $\Lambda^{SM}$ is larger than $\tau_{asc}$ which is the pre-defined association threshold. Although the object and observation are associated by the hungarian algorithm, those with affinity, smaller than $\tau_{asc}$, are ignored. So, we only calculate the appearance affinity for pairs which have $\Lambda^{SM}$ larger than $\tau_{asc}$. This can save processing time a lot.

\subsection{Scene Adaptive Detection Filtering}
In public benchmarks\cite{Taixe15, Milan16a}, detections, extracted by ACF \cite{Dollar14} or DPM \cite{Felzenszwalb10}, are given as default. Both ACF and DPM are not the state-of-the-art detectors and make a lot of false-positive, false-negative errors. So, it is necessary to filter out noisy detections for better performance. It is common to filter out noisy detections using given detection confidences which are also produced by detector. A lot of previous works simply filter out detections which have lower confidences then the pre-defined constant threshold $\tau_{const}$. However, the distribution of detection confidence is variable depending on tracking environment. We show an example in Figure 5. The average detection confidence is high in PETS09-S2L1 which is taken by a static camera and in which sizes of objects are constant. In contrast, the average confidence is low in ETH-SUNNYDAY which is taken from a dynamic camera in highly illuminated environments and in which sizes of objects are variant. If the detection threshold($\tau_{const}$) is fixed to work well in PETS09-S2L1, a lot of true-positive detections are filtered out in ETH-SUNNYDAY dataset(see Figure 5(b)). Even it varies in the same scene as time flows. So, we propose a simple method to adaptively decide the threshold depending on scene as:
\begin{equation}
\tau_t=(1-\rho^t)\tau_{sa}+\rho^t\tau_{const}
%&D = \mathcal{N}(\mu,\sigma)
\end{equation}
where $\tau_t$ is the detection threshold at the frame $t$. Detections with the confidence lower than $\tau_t$ are eliminated before tracking in the frame-$t$. The first term of the right-hand side of the equation is a scene adpative threshold($\tau_{sa}$) which considers inter-scene, intra-scene difference(described in Figure 5(a)). $\tau_{sa}$ is defined as:
\begin{equation}
\begin{aligned}
\tau_{sa}&=\underset{\tau}{\arg\min}\Vert(\beta P(D^{10}_t\leq\tau)\\&\qquad+(1-\beta)P(D^{all}_t\leq\tau))-p_d \Vert^2
\end{aligned}
\end{equation}
Two cumulative distribution functions($P(D\leq\tau)$) of the gaussian variable($D$) are combined through $\beta$. The gaussian variable($D=\mathcal{N}(\mu, \sigma)$) is derived from the average($\mu$) and standard deviation($\sigma$) of detection confidences. $D^{10}_t$ is calculated by detection confidences of recent 10 frames. Because of the reason that $D^{10}_t$ is usually calculated by the small number of samples, $D^{all}_t$, calculated by all detection confidences collected until the current frame, is needed for smoothness. $\beta$ controls the degree of smoothness and $p_d$ is an important constant which decide $\tau_{sa}$. We found that 0.4 generally works best(Figure 8(b)). The second part of Eq. 9 is the pre-defined threshold($\tau_{const}$). Because our tracker operates in fully online way, this pre-defined threshold is needed for first a few frames when the number of detection samples is small to calculate distribution($\tau_{sa}$). Its proportion($\rho^t$) gets smaller as frames($t$) gets bigger. We heuristically selected $\rho$ as 0.95.

\begin{figure}[t]
\centering
\begin{subfigure}[t]{.25\textwidth}
\centering
\includegraphics[width=1\linewidth]{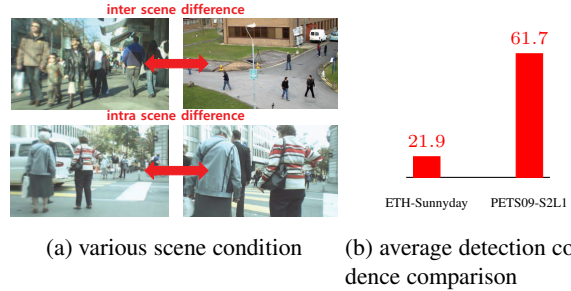}
\subcaption{various scene condition}
\end{subfigure}
\begin{subfigure}[t]{.2\textwidth}
\centering
\begin{tikzpicture}[font=\scriptsize]
\pgfplotsset{width=1\linewidth,height=3.5cm}
\begin{axis}[
hide y axis,
axis x line*=bottom,
	ybar,
    enlargelimits=0.2,
    xtick={1, 2},
    x tick style={draw=none},
    xticklabels={\tiny{ETH-Sunnyday}, \tiny{PETS09-S2L1}},
    ylabel={\#participants},
    nodes near coords,
    nodes near coords align={vertical},
    every axis plot/.append style={ybar, bar width=10, fill}
    ]
\addplot[red] coordinates {(1, 21.9) (2, 61.7)};
\end{axis}
\end{tikzpicture}
\subcaption{average detection confidence comparison}
\end{subfigure}
   \caption{(a)Upper row shows different scene condition between ETH-Sunnyday and PETS09-S2L1. Lower row shows the varying scene condition between different frames of ETH-Pedcross2. (b)Comparison of average detection confidence between two scenes(ETH-Sunnyday, PETS09-S2L1)}
\label{fig:long}
\label{fig:onecol}
\end{figure}

\section{Experiments}

In this section, we explain detail of implementation and show the improvement in tracking performance by attaching our methods one-by-one. Also, we compare the performance of our tracker with other public trackers. 

\begin{figure}[t]
\begin{center}
   \includegraphics[width=0.9\linewidth]{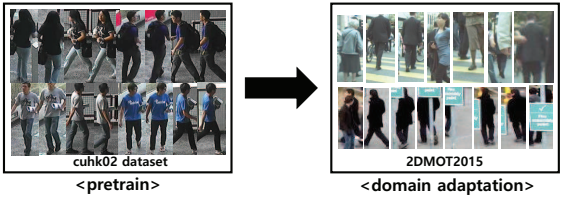}
\end{center}
\vspace{-5mm}
   \caption{Our 2-step training process. CUHK02 dataset is clear so is good to learn general concept of similarity. In contrast, 2DMOT2015 contains occluded, noisy samples. Its good to learn real-world tracking situation.}
\label{fig:long}
\label{fig:onecol}
\end{figure}

\begin{figure}[t]

\centering
\begin{subfigure}{.2\textwidth}
\centering
\begin{tikzpicture}[font=\scriptsize]
\pgfplotsset{width=1.4\linewidth,height=3cm}
\begin{axis}[
hide y axis,
axis x line*=bottom,
	ybar,
    enlargelimits=0.2,
    xtick={1, 2, 3},
    x tick style={draw=none},
    xticklabels={color hist, w/o ham, ham},
    ylabel={\#participants},
    nodes near coords,
    nodes near coords align={vertical},
    every axis plot/.append style={ybar, bar width=15, fill}
    ]
\addplot[red] coordinates {(1, 30.1) (2, 32.0) (3, 32.6)};
\end{axis}
\end{tikzpicture}
\subcaption{2DMOT2015 training}
\end{subfigure}%
\qquad
\begin{subfigure}{.2\textwidth}
\centering
\begin{tikzpicture}[font=\scriptsize]
\pgfplotsset{width=1.5\linewidth,height=3cm}
\begin{axis}[
hide y axis,
axis x line*=bottom,
	ybar,
    enlargelimits=0.2,
    xtick={1, 2, 3, 4},
    x tick style={draw=none},
    xticklabels={color hist, cuhk, 2015train, 2-step},
    ylabel={\#participants},
    nodes near coords,
    nodes near coords align={vertical},
    every axis plot/.append style={ybar, bar width=15, fill}
    ]
\addplot[blue] coordinates {(1, 31.7) (2, 32.5) (3, 32.6) (4, 33.0)};
\end{axis}
\end{tikzpicture}
\subcaption{MOT16 training}
\end{subfigure}
\caption{MOTA improvement by applying our methods. We tested the contribution of our methods on two different training dataset. In each graph, left-most bar shows the result using color-histogram based appearance model as mentioned in Eq. 3. (a)2DMOT2015 training set. The middle bar shows that our joint-input siamese network outperforms color histogram based appearance model without HAM. Right-most bar shows that performance is improved by using historical appearance matching. (b)MOT16 training set. Second and third bars from left show the result from network trained by single dataset, CUHK or 2015train. The right-most bar verify the outperforming accuracy from 2-step training}
%\label{fig:long}
%\label{fig:onecol}
\end{figure}
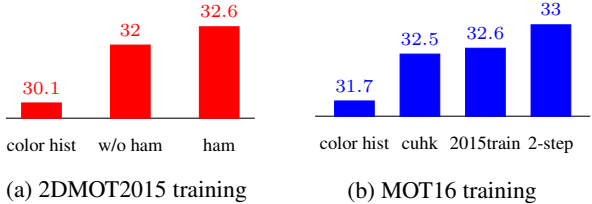

\subsection{Two-step siamese network training}

A Network may be confused if it learns directly from tracking sequences. As described in Figure 6(2DMOT2015), there exist a lot of occluded or noisy objects which are marked as ground-truth objects. If a network trains from those samples, it may decrease the performance. For this reason, it would be better to train the network about general concept of appearance comparison before training it with examples from real tracking sequences. We separated training process into two steps: pre-train from CUHK02 dataset and domain adaptation from tracking sequences. CUHK02 \cite{li2013} was developed for the person re-identification task and contains 1816 identities each of which has 4 different samples. All images are clear and not occluded. So, this is proper dataset to learn general concept of appearance comparison. First, we trained our network with learning rate $10^{-3}$ for 300 epochs(training converges). In each epoch, 3000 pairs, positive:negative ratio 1:1, are trained with mini batch size 100. After pre-train, we decreased learning rate to $10^{-4}$ and re-train it from 2DMOT2015 training sequences in a similar way as pre-training step. Cross-entropy loss and stochastic gradient descent(SGD) optimizer are used for back-propagation.

\iffalse
\subsection{Evaluation metric}
It is common to use Multi Object-Tracking Accuracy(MOTA) to evaluate overall performance of tracker. MOTA is calculated as : 
\begin{equation*} 
MOTA = 1-\frac{\sum_t{(FN_t + FP_t + IDS_t)}}{\sum_t{GT_t}}\\ 
\end{equation*}
It considers false positive, false negative, ID-switching simultaneously. Additionally we also see the IDF1 metric to compensate the limitation of IDS. Although correct label re-identified wrongly labeled object, IDS increases. Instead, IDF1 considers the overall consistency of tracklet.
\fi
\subsection{Performance evaluation}
\iffalse
\textbf{Joint-input siamese network}: We compared the accuracy of our network with \cite{Bae18}. It used cost function based siamese network taking two inputs trained from CUHK02 dataset. In Figure 7, our network shows accuracy which outperforms \cite{Bae18}. We guess this is because of two concatenated images. Cost function based network infers independently from each input image. But our network can take information from two images simultaneously through whole network.
\fi

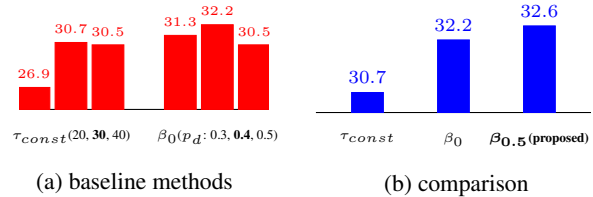
\begin{figure}[t]
\centering
\begin{subfigure}{.2\textwidth}
\centering
\begin{tikzpicture}[font=\scriptsize]
\pgfplotsset{width=1.4\linewidth,height=3cm}
\begin{axis}[
hide y axis,
axis x line*=bottom,
	ybar,
	%axis y line=center,
    enlargelimits=0.35,
    xtick={1, 2},
    x tick style={draw=none},
    %y tick style={draw=none},
    xticklabels={\tiny{$\tau_{const}$({20, \textbf{30}, 40})}, \tiny{$\beta_0$({$p_d$: 0.3, \textbf{0.4}, 0.5})}},
    %yticklabels={,,},
    %axis on top=true,
    %ylabel={\#participants},
    nodes near coords,
    nodes near coords align={vertical},
    every node near coord/.append style={font=\tiny},
    every axis plot/.append style={ybar, bar width=12, fill}
    ]
\addplot[red] coordinates {(1, 26.9) (2, 31.3)};
\addplot[red] coordinates {(1, 30.7) (2, 32.2)};
\addplot[red] coordinates {(1, 30.5) (2, 30.5)};
\end{axis}
\end{tikzpicture}
\subcaption{baseline methods}
\end{subfigure}%
\quad
\begin{subfigure}{.25\textwidth}
\centering
\begin{tikzpicture}[font=\tiny]
\pgfplotsset{width=1.2\linewidth,height=3cm}
\begin{axis}[
%hide y axis,
axis x line*=bottom,
	ybar,
    enlargelimits=0.3,
    xtick={1, 2, 3},
    x tick style={draw=none},
    xticklabels={\tiny{$\tau_{const}$}, \tiny{$\beta_0$}, $\boldsymbol{\beta_{0.5}}$\textbf{(proposed)}},
    ylabel={},
    hide y axis,
    nodes near coords,
    nodes near coords align={vertical},
    every node near coord/.append style={font=\scriptsize},
    every axis plot/.append style={ybar, bar width=12, fill}
    ]
\addplot[blue] coordinates {(1, 30.7) (2, 32.2) (3, 32.6)};
\end{axis}
\end{tikzpicture}
\subcaption{comparison}
\end{subfigure}
\caption{(a)MOTA scores in 2DMOT2015 training-set from two different kinds of detection threshold. $\tau_{const}$: pre-defined threshold(20, 30, 40 from left to right). $\beta_0$: special case of Eq. 10 when $\beta=0$. This means that it doesn't consider intra scene difference. (b)MOTA score comparison between proposed method($\beta = 0.5$) and other baseline methods($\tau_{const}$, $\beta_0$). We chose the best score of each method in (a).}
\end{figure}

\textbf{Performance improvement}: To prove the contribution of our methods in tracking performance, we provide several experimental results(Figure 7 and 8). In Figure 7, we compared results by sequentially attaching our method related to appearance affinity. We tested on two kinds of training dataset(2DMOT2015, MOT16). In 2DMOT2015 dataset, we tested the validity of our historical appearance matching method. For the fair experiment, we used the network pre-trained from CUHK02 dataset without additional training from 2015 training-set. As you can see in Figure 7(a), It is clear that the historical appearance matching improve the overall tracking performance. In MOT16 training-set, we tested validity of the 2-step training method. As you can see in Figure 7(b), the 2-step training method shows the highest MOTA score outperforming networks which are trained from each single dataset. In Figure 8, you can find experimental results which prove necessity of scene adaptive detection filtering. We compared our MOTA score with scores from other filtering methods($\tau_{const}$, $\tau_t(\beta_0)$).

\textbf{Comparison with other trackers}: We compared our methods with several trackers on 2DMOT2015 and MOT17 benchmark. Because ID consistency metrics(IDSw, IDF1) are critically affected by Fasle-Positive(FP) and False-Negative(FN), we carefully selected other trackers, having similar number of FP and FN with ours, to be compared. Overall comparison result is on Table 1 and Table 2. In Table 1, we provide results in 2DMOT2015 benchmark. We measured our tracker in two different ways, SADF and VisBest. In SADF, we applied our proposed scene adaptive detection filtering method($p_d=0.3$). Although SADF doesn't show the state-of-the-art MOTA, it shows better performance than methods which don't use appearance reasoning \cite{Matilla16, Wang16}. This proves the effectiveness of deep appearance feature. Also, SADF shows better ID-consistency(IDF1, IDSw) than fine-tuned baseline method \cite{Taixe16}. IDF1 was proposed to compensate the limitation of IDSw metric. High performance in both IDF1 and IDSw metrics proves that our tracker can manage object ID consistently. In VisBest, we heuristically chose a different detection filtering threshold, visually seems best, for each sequence. VisBest removed FN a lot and produced near state-of-the-art MOTA, highest IDF1 and second-best IDSw except SADF. It is remarkable that ours showed far better ID-consistency than that of \cite{Taixe16, Bae18} which used a similar siamese network. We guess it attributes to our historical appearance matching and two-step training method. In Table 2, we compared with a few state-of-the-art trackers in MOT17 benchmark. Ours shows competitive performance in all metrics. Even our tracker outperforms state-of-the-art LSTM based tracker \cite{Kim2018} in MOTA and IDSw metrics.

\begin{table}[t]
\captionsetup{font=small}
 \begin{center}
 {\footnotesize
\begin{tabular}{|l|ccccc|}
\hline
\multicolumn{1}{|c|}{Method} &MOTA$\uparrow$ &IDF1$\uparrow$ &IDSw$\downarrow$ &FP$\downarrow$ &FN$\downarrow$\\
\hline
\multicolumn{6}{|c|}{offline}\\
\hline
SiameseCNN \cite{Taixe16} & 29.0 & 34.3 & 639 & \textcolor{blue}{\textbf{5160}} & 37798\\
LP\_SSVM \cite{Wang16} & 25.2 & 34.0 & 646 & 8369 & 36932\\
\hline
\multicolumn{6}{|c|}{online}\\
\hline
CDA\_DDAL \cite{Bae18} & \textcolor{red}{\textbf{32.8}} & 38.8 & 614 & \textcolor{red}{\textbf{4983}} & \textcolor{blue}{\textbf{35690}}\\
PHD\_GSDL \cite{Fu2018} & \textcolor{blue}{\textbf{30.5}} & 38.8 & 879 & 6534 & \textcolor{red}{\textbf{35284}}\\
SCEA \cite{Yoon16} & 29.1 & 37.2 & 604 & 6060 & 36912\\
oICF \cite{Keritz16} & 27.1 & \textcolor{blue}{\textbf{40.5}} & \textcolor{blue}{\textbf{454}} & 7594 & 36757\\
EAMTT \cite{Matilla16} & 22.3 & 32.8 & 833 & 7924 & 38982\\
\hline
Ours(VisBest) & 28.6 & \textcolor{red}{\textbf{41.4}} & 460 & 7485 & 35910\\
Ours(SADF) & 25.2 & 37.8 & \textcolor{red}{\textbf{357}} & 7330 & 38275\\
\hline
\end{tabular}}
\end{center}
\vspace{-5mm}
\caption{Comparison in 2DMOT2015 benchmark. Best(red) and runner-up(blue) scores in table are marked in bold.}
\end{table}

\begin{table}[t]
\captionsetup{font=small}
 \begin{center}
 {\footnotesize
\begin{tabular}{|l|ccccc|}
\hline
\multicolumn{1}{|c|}{Method} &MOTA$\uparrow$ &IDF1$\uparrow$ &IDSw$\downarrow$ &FP$\downarrow$ &FN$\downarrow$\\
\hline
\multicolumn{6}{|c|}{offline}\\
\hline
{\scriptsize MHT\_DAM} \cite{Kim2015} & \textcolor{red}{\textbf{50.7}} & 47.2 & 2314 & \textcolor{blue}{\textbf{22875}} & \textcolor{red}{\textbf{252889}}\\
{\scriptsize MHT\_bLSTM} \cite{Kim2018} & 47.5 & \textcolor{red}{\textbf{51.9}} & \textcolor{blue}{\textbf{2069}} & 25981 & 268042\\
\hline
\multicolumn{6}{|c|}{online}\\
\hline
PHD\_GSDL \cite{Fu2018}  & 48.0 & 49.6 & 3998 & 23199 & \textcolor{blue}{\textbf{265954}}\\
EAMTT \cite{Matilla16} & 42.6 & 41.8 & 4488 & 30711 & 288474\\
\hline
Ours(SADF) & \textcolor{blue}{\textbf{48.3}} & \textcolor{blue}{\textbf{51.1}} & \textcolor{red}{\textbf{1871}} & \textcolor{red}{\textbf{20967}} & 269038\\
\hline
\end{tabular}}
\end{center}
\vspace{-5mm}
\caption{Comparison in MOT17 benchmark. We applied $\tau_{sa}$($p_d=0.3$) for DPM and didn't apply threshold for FRCNN and SDP.}
\end{table}

\section{Conclusion}
We proposed several methods to overcome temporal errors which occur because of occulsion and noisy detections. First, we designed the joint-input siamese network for appearance matching and trained it using the 2-step training method. And we applied historical appearance matching method to break ambiguity. Finally, we tried to find an adaptive detection threshold which generally works well in all sequences. As confirmed in experiment, our tracker showed improved performance, especially in ID consistency metrics. But there is a limitation of our work. Our network only takes cropped patches as input and lacks contextual information. In our future work, we will try to exploit contextual information instead of directly cropping patches from image.

\section{Acknowledgement}
This work was financially supported by the ICT R\&D program of MSIP/IITP [2014-0-00077, Development of global multi-target tracking and event prediction techniques based on real-time large-scale video analysis] and Lotte Data Communication Company.

{\small
\bibliographystyle{ieee}
\bibliography{egbib}
}

\end{document}